\documentclass[preprint,12pt]{elsarticle}

\usepackage[utf8]{inputenc}
\usepackage[T1]{fontenc}
\usepackage{lmodern}
\usepackage{dirtree}
\usepackage{graphicx}
\usepackage{xcolor}
\usepackage{amsmath}
\usepackage{placeins}
\usepackage{subcaption}   
\usepackage{float}
\usepackage{geometry}
\usepackage{longtable}
\usepackage{array}
\usepackage{booktabs}
\usepackage{enumitem}
\usepackage{xurl}
\usepackage{hyperref}
\setlist[itemize]{left=0pt,label=--,itemsep=2pt,topsep=2pt}

\journal{Data in Brief}

\begin{document}

\begin{frontmatter}



\title{A labeled dataset of simulated phlebotomy procedures for medical AI: polygon annotations for object detection and human--object interaction}

\author[inst1]{Raúl Jiménez Cruz}

\author[inst1]{César Torres-Huitzil}

\author[inst2]{Marco Franceschetti\corref{cor1}}
\cortext[cor1]{Corresponding author}
\ead{marco.franceschetti@unisg.ch}

\author[inst2]{Ronny Seiger}

\author[inst1]{Luciano García-Bañuelos}

\author[inst2]{Barbara Weber}

\affiliation[inst1]{organization={Tecnológico de Monterrey, Escuela de Ingeniería y Ciencias},
            addressline={Av. Eugenio Garza Sada 2501 Sur}, 
            city={Monterrey, N.L.},
            postcode={64849}, 
            country={ México}}

\affiliation[inst2]{organization={Institute of Computer Science},
            addressline={University of St.Gallen, Rosenbergstrasse 30}, 
            city={St.Gallen},
            postcode={9000}, 
            country={Switzerland}}


\begin{abstract}
This data article presents a dataset of 11{,}884 labeled images documenting a simulated blood extraction (phlebotomy) procedure performed on a training arm. Images were extracted from high-definition videos recorded under controlled conditions and curated to reduce redundancy using Structural Similarity Index Measure (SSIM) filtering. An automated face-anonymization step was applied to all videos prior to frame selection. Each image contains polygon annotations for five medically relevant classes: \emph{syringe}, \emph{rubber band}, \emph{disinfectant wipe}, \emph{gloves}, and \emph{training arm}. 
The annotations were exported in a segmentation format compatible with modern object detection frameworks (e.g., YOLOv8), ensuring broad usability.
This dataset is partitioned into training (70\%), validation (15\%), and test (15\%) subsets and is designed to advance research in medical training automation and human–object interaction. It enables multiple applications, including phlebotomy tool detection, procedural step recognition, workflow analysis, conformance checking, and the development of educational systems that provide structured feedback to medical trainees.
The data and accompanying label files are publicly available on Zenodo.
\end{abstract}


\end{frontmatter}

\pagebreak
\section*{Specifications Table}

\renewcommand{\arraystretch}{1.2}
\begin{longtable}{p{4.2cm} p{11.8cm}}
\toprule
\textbf{Subject} & Computer Science / Artificial Intelligence \\
\midrule
\textbf{Specific subject area} & Computer vision for medical training; human--object interaction (HOI) \\
\midrule
\textbf{Type of data} & Images (JPG, variable resolution; longer side 640 px with aspect ratio preserved, e.g., 640$\times$480, 640$\times$360); Annotation files (TXT, YOLOv8 polygon segmentation~\cite{ref:yolo}); Figure(s); README \\
\midrule
\textbf{How data were acquired} & High-definition video recordings (1920$\times$1080 @30\,FPS) of simulated phlebotomy on a training arm; static tripod camera; controlled lighting. Frames selected with SSIM-based filtering~\cite{ref:ssim}; face anonymization with Python \texttt{face\_recognition}~\cite{ref:face} + OpenCV~\cite{ref:opencv}. \\
\midrule
\textbf{Data source location} & Institute of Computer Science, University of St.Gallen, Switzerland \\
\midrule
\textbf{Data accessibility} & Repository: Zenodo~\cite{jimenez_cruz_2025_16924786} \\
 & DOI / PID: \texttt{10.5281/zenodo.16924786} \\
 & Direct URL: \url{https://zenodo.org/records/16924786} \\
 & Access instructions: Public, CC BY 4.0 \\
\midrule
\textbf{Related research article} & None. \\
\bottomrule
\end{longtable}

\pagebreak
\section*{Value of the Data}
\begin{figure}[H]
\centering
\includegraphics[width=0.8\textwidth]{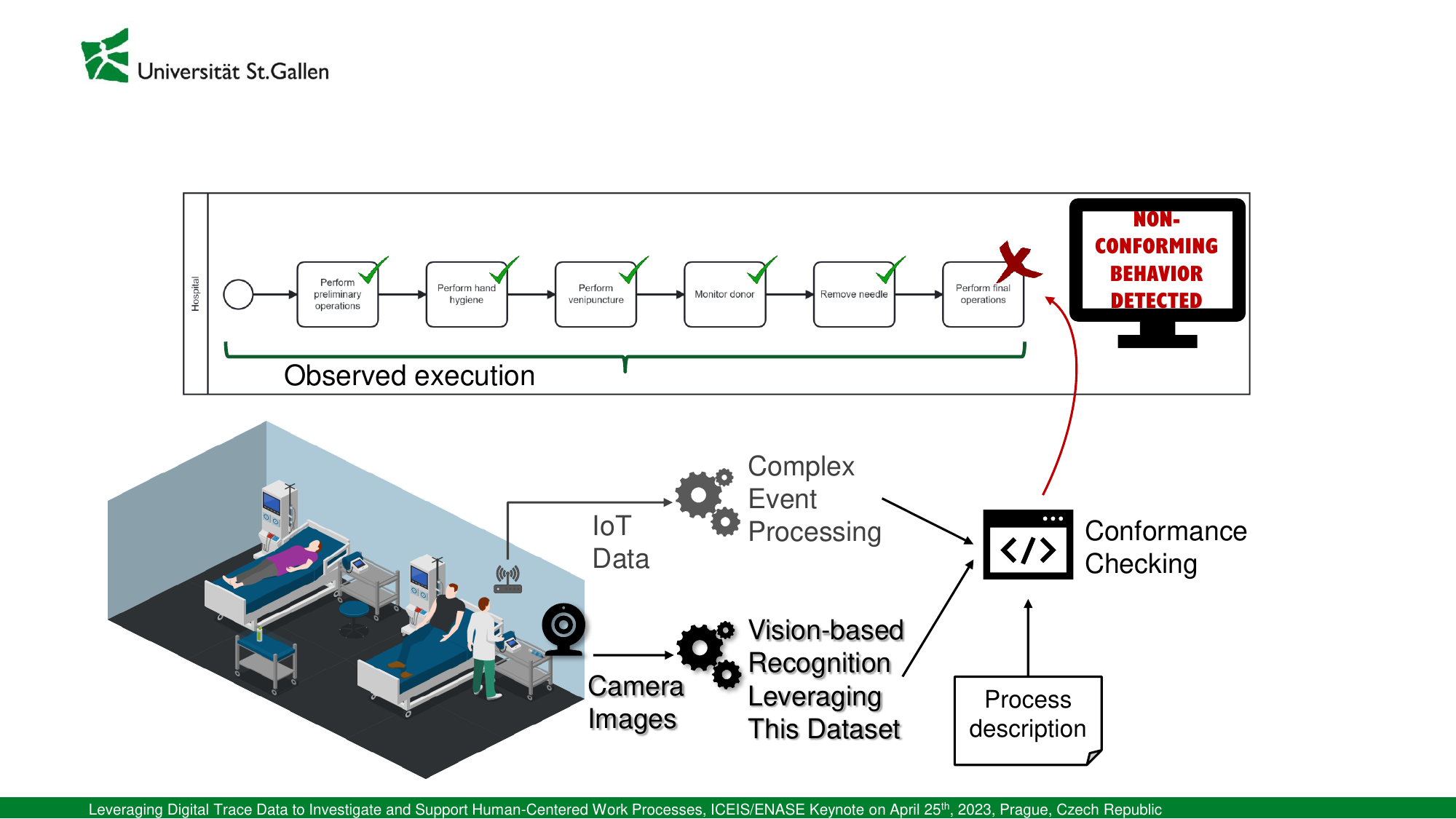}
\caption{Vision-based conformance checking of a phlebotomy process, combined with IoT data~\cite{seiger2026online}.}
\label{fig:conformance_checking}
\end{figure}

\begin{itemize}
    \item Provides a curated dataset with polygon segmentation for five key objects involved in simulated phlebotomy, enabling fine-grained spatial reasoning beyond bounding boxes.
    \item Facilitates research on human--object interaction, tool detection, and step recognition in clinical skill training, with splits aligned to typical machine learning workflows.
    \item Ready-to-use YOLOv8~\cite{ref:yolo} segmentation format lowers entry barriers for benchmarking object detection/segmentation models in medical education contexts.
    \item Supports applications in AR/VR educational interfaces, intelligent tutoring systems requiring reliable tool/step detection, and medical workflow analysis.
    \item Enables vision-based process conformance and compliance checking, and the provision of visual feedback based on the checking results; combined for example with IoT data (cf. Figure~\ref{fig:conformance_checking}).
\end{itemize}

\section*{Background}

Correctly executing phlebotomy--the process of drawing blood--is critical to ensure patient safety~\cite{ref:who_blood}. Mastering phlebotomy requires using specific tools in the correct sequence. As opportunities to practice under expert supervision and receive feedback are limited, automated feedback approaches provide valuable support. However, public datasets capturing realistic scenarios with precise annotations remain scarce, hindering the development of such approaches. To address this gap, this dataset provides high-quality, polygon-level annotations of camera images depicting five core objects in a simulated phlebotomy. 

\begin{figure}
\centering
\includegraphics[width=0.8\textwidth]{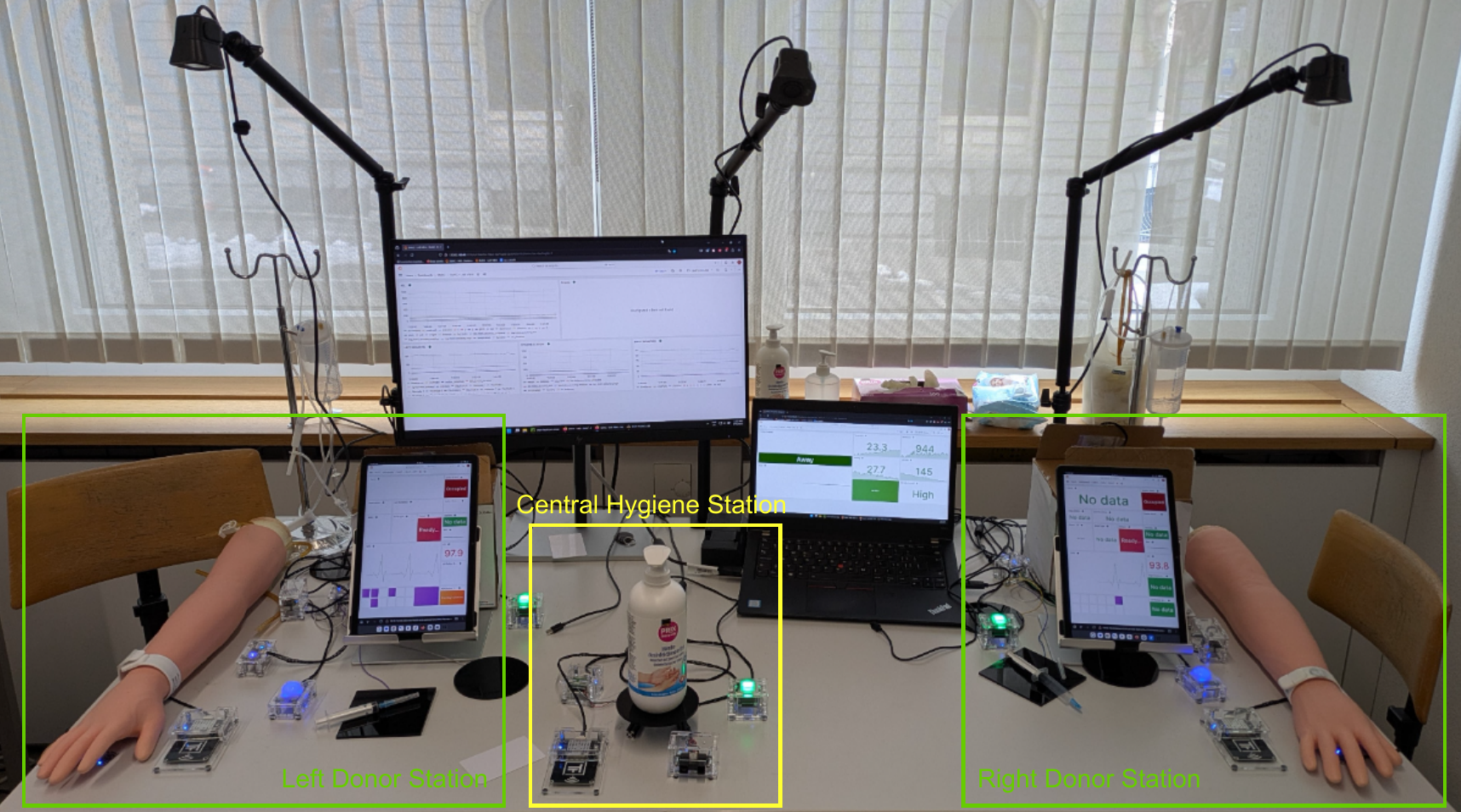}
\caption{Phlebotomy simulation and monitoring setup using IoT sensors and cameras.}
\label{fig:setup}
\end{figure}

Figure~\ref{fig:setup} illustrates a training environment providing automated feedback. It involves IoT sensors (e.g.,~proximity, motion, NFC) monitoring activities performed by healthcare workers~\cite{franceschetti2023proambition}. To cope with \emph{ambiguities} in sensor-based activity detection~\cite{franceschetti2023characterisation}, the environment is augmented with cameras detecting objects and human--object interactions.

The curation emphasizes visual diversity at the frame level, privacy preservation, and compatibility with widely adopted model formats. By documenting both data structure and acquisition methods, the article aims to support reuse across research tasks such as phlebotomy tool detection, procedural step recognition, workflow analysis, conformance checking, and the development of educational systems that provide structured feedback to medical trainees. This dataset contributes to an underexplored area, as polygon-level annotations of simulated phlebotomy are rarely reported in publicly accessible repositories.

\begin{table}[H]
\centering
\begin{tabular}{ll}
\toprule
\textbf{Class ID} & \textbf{Label name} \\
\midrule
0 & Disinfectant wipe \\
1 & Gloves \\
2 & Rubber band \\
3 & Syringe \\
4 & Training arm \\
\bottomrule
\end{tabular}
\caption{Object classes and label names used in the dataset.}
\label{tab:classes}
\end{table}

\FloatBarrier

\section*{Data Description}
\begin{figure}[H]
\centering
\begin{subfigure}{0.485\textwidth}
  \centering
  \includegraphics[width=\linewidth]{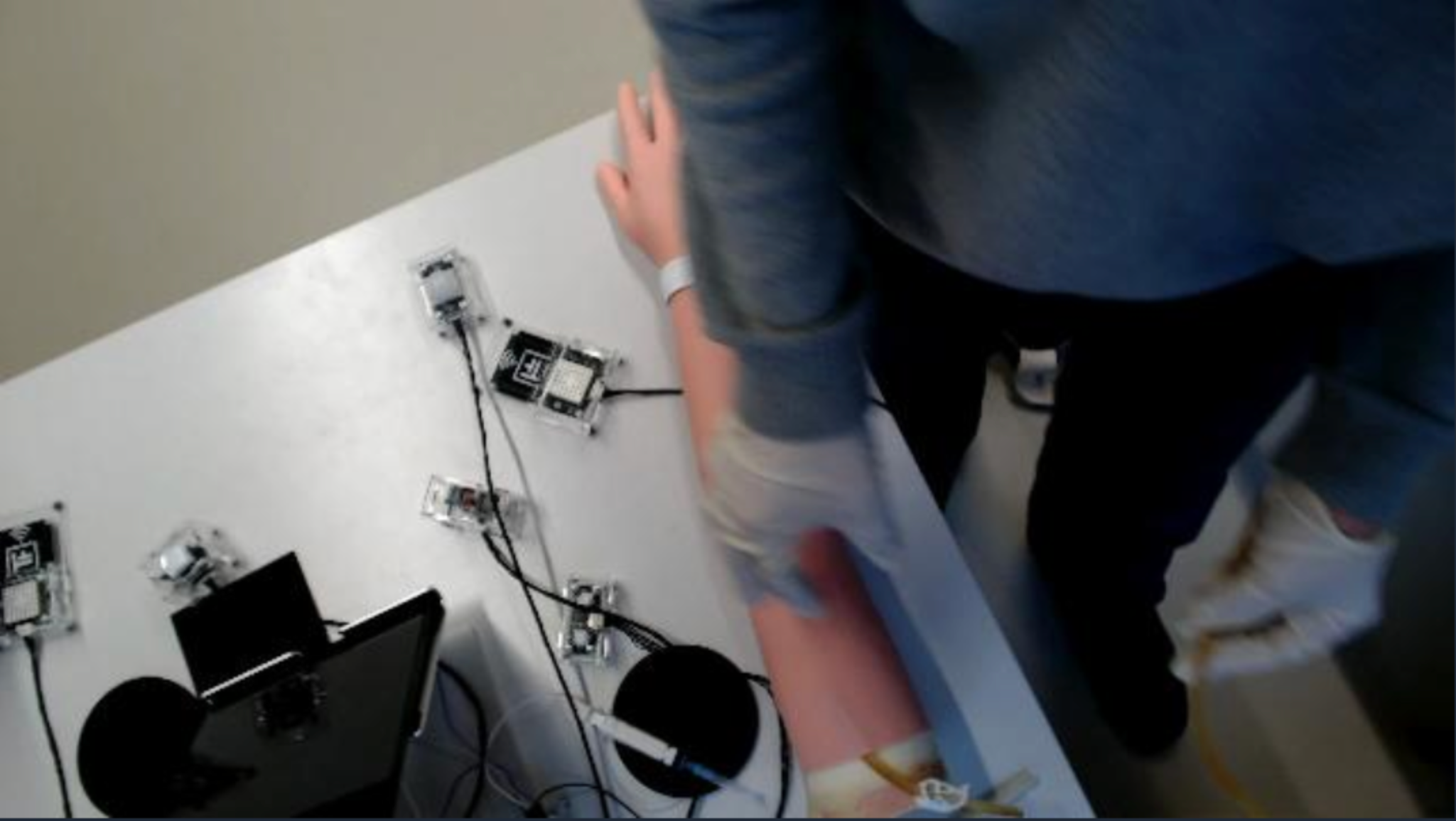}
  \caption{Original frame (1920$\times$1080).}
\end{subfigure}
\hfill
\begin{subfigure}{0.485\textwidth}
  \centering
  \includegraphics[width=\linewidth]{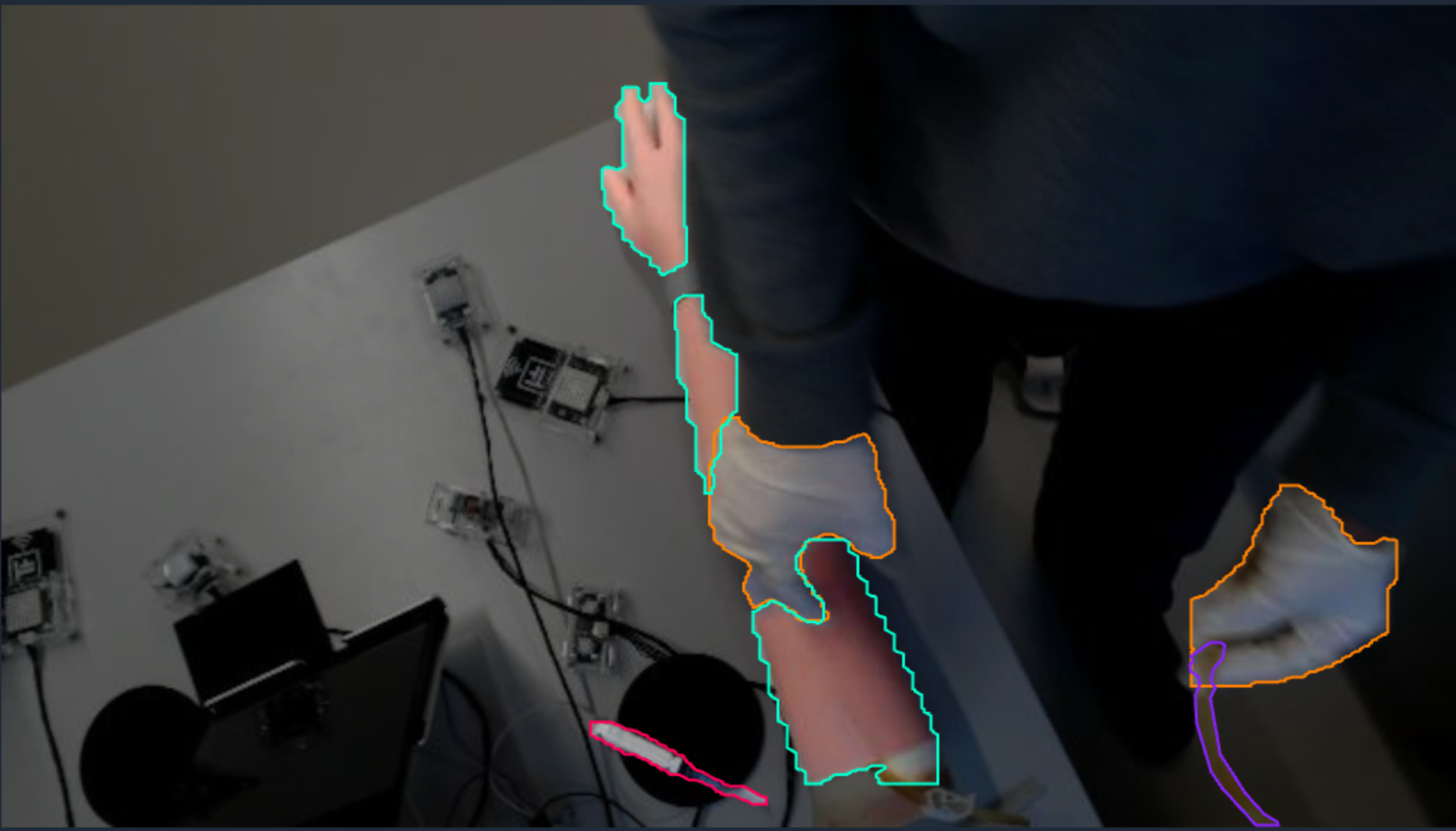}
  \caption{Annotated frame (longer side 640 px).}
\end{subfigure}
\caption{Side-by-side example of original and annotated frames showing the object classes.}
\label{fig:examples}
\end{figure}

\begin{figure}[H]
\centering
\includegraphics[width=0.85\textwidth]{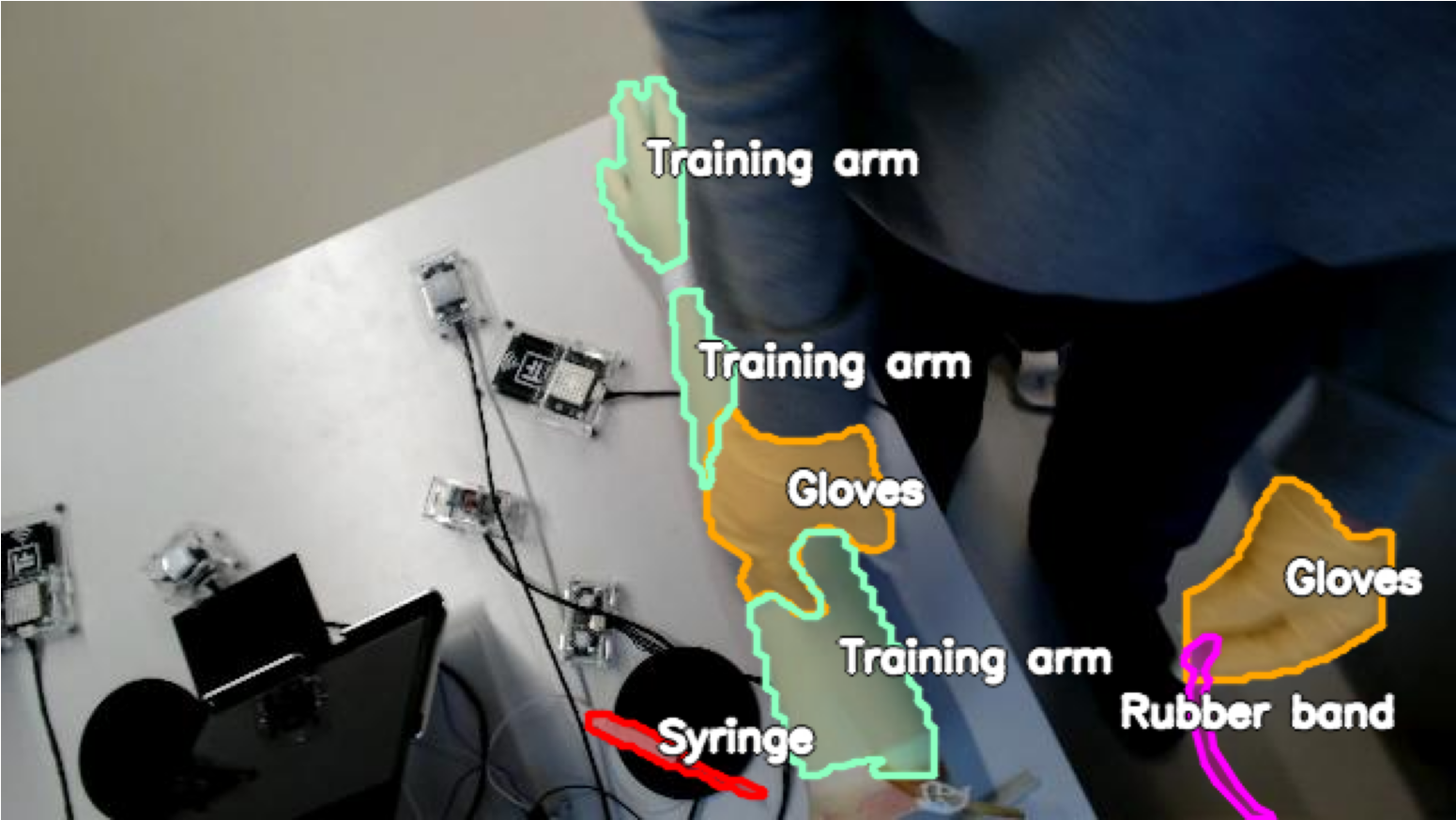}
\caption{Detailed annotated example with polygons and text labels (syringe, gloves, rubber band, training arm).}
\label{fig:example_detailed}
\end{figure}

The dataset comprises \textbf{11{,}884} RGB images (JPG, resized so that the longer side equals 640 px; the shorter side is either 480 px or 360 px, depending on the original aspect ratio adjustment) and corresponding YOLOv8-compatible polygon label files (TXT). Each label file stores class IDs and normalized polygon coordinates. Five classes are included: (0) disinfectant wipe, (1) gloves, (2) rubber band, (3) syringe, (4) training arm (cf.~Table~\ref{tab:classes}). These classes correspond to all the relevant objects in the phlebotomy process that do not autonomously provide execution data and cannot be augmented with IoT sensors providing data for reliable monitoring. In contrast, other tools--such as blood drawing machines--either record their operations automatically or can easily be augmented with IoT sensors to support process monitoring~\cite{levy2010development}. Therefore, these five visually distinctive objects capture the essential human-object interactions in the specific process while excluding other objects already monitored through alternative modalities. There are no other relevant medical instruments or objects involved in the phlebotomy process that require monitoring.
Representative examples of original and annotated frames are shown in Figure~\ref{fig:examples}. 
A more detailed annotated example, with polygons and text labels clearly visible, is shown in Figure~\ref{fig:example_detailed}. This figure highlights how object boundaries and class identifiers were represented in the dataset.

\FloatBarrier

\FloatBarrier

\paragraph{File formats}
\begin{itemize}
  \item \textbf{Images:} JPG, variable resolution with longer side 640 px (aspect ratio preserved; e.g., 640$\times$480, 640$\times$360), RGB.
  \item \textbf{Labels (YOLOv8 segmentation):} one line per instance: \texttt{class x\_1 y\_1 x\_2 y\_2 ... x\_n y\_n}, coordinates normalized to $[0,1]$.
  \item \textbf{Metadata:} \texttt{data.yaml} with class names and split paths.
\end{itemize}

\paragraph{Repository structure} The repository is structured as follows:

\dirtree{%
.1 /.
.2 README.md.
.2 data.yaml \DTcomment{YOLOv8 config (paths, names)}.
.2 train/ \DTcomment{JPG files (70\%)}.
.3 images/.
.3 labels/.
.2 val/ \DTcomment{JPG files (15\%)}.
.3 images/.
.3 labels/.
.2 test/ \DTcomment{JPG files (15\%)}.
.3 images/.
.3 labels/.
}

\section*{Experimental Design, Materials and Methods}

\paragraph{Acquisition setup}
Videos were recorded with a static tripod camera (1920$\times$1080 @30\,FPS) under controlled lighting at the left donor station (cf.~Figure~\ref{fig:setup}). Recordings were conducted under varying ambient lighting conditions (daylight and mixed artificial light) and from two slightly different fixed camera angles, introducing moderate variations in brightness, perspective, and occasional partial occlusions of the trainee’s hands and tools. Each session followed a standardized simulated phlebotomy process on a training arm, using real instruments (rubber band, disinfectant wipe, syringe, gloves). The process was validated with domain experts from the Cantonal Hospital of St.Gallen. The data were collected during the execution of the process by Bachelor and PhD students of computer science with prior instruction. Together with data from IoT sensors (cf.~\cite{kurz_2025_15436046}), the detected objects and human--object interactions are used to verify the execution of the phlebotomy process for conformance and compliance~\cite{franceschetti2025proambition}.

\paragraph{Privacy and preprocessing}
Before any frame extraction, we applied automatic face detection and blurring using Python \texttt{face\_recognition} and OpenCV (Gaussian blur). Frame selection used SSIM-based filtering~\cite{ref:ssim} to reduce redundancy: every $k$th frame was compared to the previous kept frame, saving only frames with SSIM $< 0.95$ (example threshold).

\paragraph{Frame curation and standardization}
Filenames were standardized and splits were organized into unified \texttt{train/val/test} folders. Square padding was not applied in the released data (padding can be introduced at training time if required by specific frameworks).

\paragraph{Annotation pipeline}
Polygon annotations were produced in Roboflow \cite{ref:roboflow}, an online platform that facilitates image annotation, dataset management, and quick training/validation of computer vision models. A \emph{golden set} of 8,743 diverse images was manually labeled by an expert and used to bootstrap a semi-automatic pipeline:
\begin{enumerate}
  \item Train a provisional segmentation model on the golden set.
  \item Auto-label the remaining images.
  \item Human verification/correction of all auto-generated masks for class integrity and boundary precision.
\end{enumerate}

Annotation followed a model-assisted workflow using the Roboflow 3.0 Instance Segmentation (Fast) configuration under default training parameters. An initial small golden set was used to train a preliminary model but was later replaced by a larger, higher-quality golden set of 8,743 manually labeled images. The retrained model then assisted in labeling the remaining frames, reaching a total of 11,884 annotated images. All assisted annotations were manually corrected by the main annotator, and random samples were cross-checked by a second annotator to ensure consistency. Exports follow a format compatible with YOLOv8 segmentation (TXT), but can also be converted for other frameworks.

\begin{figure}[htbp]
\centering
\includegraphics[width=\textwidth]{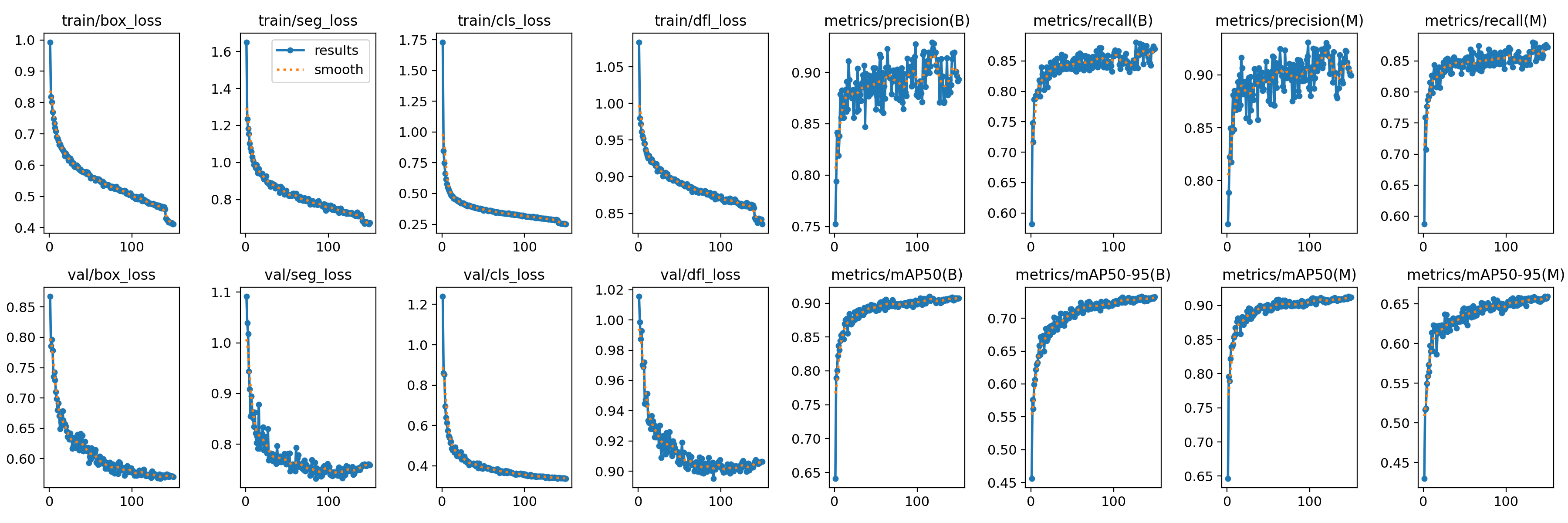}
\caption{Training and validation curves over 150 epochs for YOLOv8 segmentation. 
Loss terms: \emph{box\_loss} (box regression), \emph{cls\_loss} (classification), \emph{seg\_loss} (segmentation masks), and \emph{dfl\_loss} (distribution focal loss for box refinement). 
Metrics: precision, recall, and mean Average Precision at IoU thresholds; mAP50 corresponds to IoU = 0.50 and mAP50--95 averages across 0.50:0.05:0.95.}
\label{fig:training}
\end{figure}

\paragraph{Quality assurance and validation}
Annotation quality was ensured through an iterative combination of model-assisted labeling, manual correction, and verification by multiple annotators, complemented by sanity-check trainings with YOLOv8 segmentation.
\begin{itemize}
  \item Visual Quality Assurance: removal of blurry/occluded frames and ensuring class presence when expected.
  \item Label Quality Assurance: verification of correct class assignment and polygon tightness (snap-to-edge tools in Roboflow).
  \item Sanity-check training: we trained YOLOv8 (segmentation) for up to 150 epochs on Apple Silicon (M1 Pro). We monitored four loss components—\emph{box\_loss} (box regression), \emph{cls\_loss} (classification), \emph{seg\_loss} (segmentation masks), and \emph{dfl\_loss} (distribution focal loss for box refinement)—and evaluation metrics: precision, recall, mAP50 (IoU = 0.50), and mAP50--95 (mean AP over 0.50:0.05:0.95). Curves stabilized at high values (see Figure~\ref{fig:training}). 
  \item Rapid annotation consistency check: Roboflow’s built-in training environment was used to train a lightweight YOLOv8 segmentation model directly on the golden set. Outputs on held-out frames confirmed that the labeled classes were separable and that no systematic misannotations occurred. This step was complementary to the local sanity-check training.
\end{itemize}

\paragraph{Code and software}
The processing and anonymization scripts were implemented in Python (OpenCV, \texttt{face\_recognition}). Annotation via Roboflow; exports in YOLOv8 format. Example \texttt{data.yaml} and environment files are provided in the repository folder.

\section*{Limitations}
The dataset captures a simulated environment with a training arm and may not fully reflect clinical variability (backgrounds, illumination, camera viewpoints, hand postures). Only five object classes are included; specialized accessories or alternative tool variants are not annotated. Temporal labels for process steps are not part of this release (frame-based only). While face anonymization was applied, residual risk in edge cases cannot be entirely excluded and should be reassessed if original videos are reused. Future versions may incorporate multiple camera angles, broader demographics of trainees, additional classes, and temporally annotated sequences. Although this dataset emphasizes visual fidelity and annotation quality, its scope is limited to simulated environments and should be complemented with real clinical data for broader generalization.

\section*{Ethics Statement}
\noindent
This work involves simulated procedures on a training arm. No real patients were recorded. Any human faces incidentally present in the videos were automatically anonymized prior to frame extraction. The authors have read and comply with the ethical requirements of Data in Brief.

\section*{CRediT Author Statement}
\noindent
Raúl Jiménez Cruz: Data curation; Annotation; Writing -- original draft. \\
César Torres-Huitzil: Data management; Supervision. \\
Marco Franceschetti: Project member; Writing -- review \& editing. \\
Ronny Seiger: Project member; Writing -- review \& editing. \\
Luciano García-Bañuelos: Project administration; Supervision. \\
Barbara Weber: Writing -- review \& editing; Project administration; Supervision.

\section*{Acknowledgements}
\noindent
This work has received funding from the Swiss National Science Foundation under Grant No. IZSTZ0\_208497 (\emph{ProAmbitIon} project).

\section*{Declaration of Competing Interests}
\noindent
The authors declare that they have no known competing financial interests or personal relationships that could have appeared to influence the work reported in this paper.

\section*{Data Availability}
\noindent
The dataset is publicly available on Zenodo~\cite{jimenez_cruz_2025_16924786}: 
DOI: \url{https://doi.org/10.5281/zenodo.16924786}. 
Access: Public (license: CC BY 4.0).


\end{document}